\documentclass[11pt,a4paper]{article}
\usepackage[hyperref]{naaclhlt2019}
\usepackage{times}
\usepackage{amsmath}
\usepackage{amssymb}
\usepackage{graphicx}
\usepackage{stackengine}
\usepackage{latexsym}
\usepackage{booktabs}
\usepackage{multirow}
\usepackage{subfig}
\usepackage{url}

\usepackage{url}

\newcommand{\ignore}[1]{}

\aclfinalcopy 

\newcommand{\parheader}[1]{{\smallskip \noindent \bf #1.}}
\newcommand{\WW}{\mathbf{W}}

\title{Progress and Tradeoffs in Neural Language Models}

\author{Raphael Tang \and Jimmy Lin\vspace{0.1cm}\\
David R. Cheriton School of Computer Science\\
University of Waterloo\vspace{0.1cm}\\
{\tt \{r33tang, jimmylin\}@uwaterloo.ca}}

\date{}

\begin{document}
\maketitle
\begin{abstract}
In recent years, we have witnessed a dramatic shift towards techniques
driven by neural networks for a variety of NLP tasks. Undoubtedly,
neural language models (NLMs) have reduced perplexity by impressive
amounts. This progress, however, comes at a substantial cost in
performance, in terms of inference latency and energy consumption,
which is particularly of concern in deployments on mobile
devices. This paper, which examines the quality--performance tradeoff
of various language modeling techniques, represents to our knowledge
the first to make this observation. We compare state-of-the-art NLMs
with ``classic'' Kneser-Ney (KN) LMs in terms of energy usage,
latency, perplexity, and prediction accuracy using two standard
benchmarks. On a Raspberry Pi, we find that orders of increase in 
latency and energy usage correspond to less change in 
perplexity, while the difference is much less pronounced on 
a desktop.

\end{abstract}

\section{Introduction}

Deep learning has unquestionably advanced the state of the
art in many natural language processing tasks, from syntactic dependency 
parsing~\cite{parser} to named-entity recognition~\cite{ner} to machine
translation~\cite{nmt}. The same certainly applies to language
modeling, where recent advances in neural language models (NLMs) have
led to dramatically better approaches as measured using standard
metrics such as perplexity~\cite{melis2017state, merity2017regularizing}.

Specifically focused on language modeling, this paper examines an
issue that to our knowledge has not been explored:\ advances in neural
language models have come at a significant cost in terms of increased
computational complexity. Computing the probability of a token
sequence using non-neural techniques requires a number of phrase
lookups and perhaps a few arithmetic operations, whereas model inference with
NLMs require large matrix multiplications consuming perhaps millions
of floating point operations (FLOPs). These performance tradeoffs are
worth discussing.

In truth, language models exist in a quality--performance tradeoff
space. As model quality increases (e.g., lower perplexity),
performance as measured in terms of energy consumption, query latency,
etc.~tends to decrease. For applications primarily running in the
cloud---say, machine translation---practitioners often solely optimize 
for the lowest perplexity. This is because such applications are 
embarrassingly parallel and hence trivial to scale in a data center 
environment.

There are, however, applications of NLMs that require less one-sided
optimizations. On mobile devices such as smartphones and tablets, for
example, NLMs may be integrated into software keyboards for next-word
prediction, allowing much faster text entry. Popular Android apps that
enthusiastically tout this technology include
SwiftKey and
Swype. The greater computational costs of NLMs lead to higher 
energy usage in model inference, translating into shorter battery 
life.

In this paper, we examine the quality--performance tradeoff in the
shift from non-neural to neural language models. In particular, we
compare Kneser--Ney smoothing, widely accepted as the state of the art
prior to NLMs, to the best NLMs today. The decrease in perplexity on
standard datasets has been well documented~\cite{melis2017state}, but to our
knowledge no one has examined the performances tradeoffs. With
deployment on a mobile device in mind, we evaluate energy usage and
inference latency on a Raspberry Pi (which shares the same ARM
architecture as nearly all smartphones today). We find that a
2.5$\times$ reduction in perplexity on PTB comes at a staggering cost in
terms of performance:\ inference with NLMs takes 49$\times$ longer and
requires 32$\times$ more energy. Furthermore, we find that impressive
reductions in perplexity translate into at best modest improvements in
next-word prediction, which is arguable a better metric for evaluating
software keyboards on a smartphone. The contribution of this paper is
the first known elucidation of this quality--performance
tradeoff. Note that we refrain from prescriptive recommendations:\ whether or 
not a tradeoff is worthwhile depends on the application. Nevertheless, 
NLP engineers should arguably keep these tradeoffs in mind when selecting a 
particular operating point.

\section{Background and Related Work}

\citet{melis2017state} evaluate recent neural language models; however, their 
focus is not on the computational footprint of each model, but rather the 
perplexity. To further reduce perplexity, many neural language model extensions 
exist, such as continuous cache pointer~\cite{cache} and mixture of 
softmaxes~\cite{yang2018breaking}. Since our focus is on comparing ``core''
neural and non-neural approaches, we disregard these extra optimizations
techniques in all of our models.

Other work focus on designing lightweight models for resource-efficient
inference on mobile devices. \citet{liu2018binarized} explore LSTMs~\cite{lstm} 
with binary weights for language modeling; \citet{botha2017natural} examine
shallow feedforward neural networks for natural language processing.

\parheader{AWD-LSTM}
\citet{merity2017regularizing} show that a simple three-layer LSTM, with proper 
regularization and optimization techniques, can achieve state of the art on 
various language modeling datasets, surpassing more complex models. 
Specifically, \citet{merity2017regularizing} apply randomized backpropagation 
through time, variational 
dropout, activation regularization, embedding dropout, and temporal activation 
regularization. A novel scheduler for optimization, non-monotonically triggered 
ASGD (NT-ASGD) is also introduced. \citet{merity2017regularizing} name their 
three-layer LSTM model trained with such tricks, ``AWD-LSTM.''

\parheader{Quasi-Recurrent Neural Networks}
Quasi-recurrent neural networks (QRNNs; \citealp{bradbury2016quasi}) achieve 
current state of the art in word-level language 
modeling~\cite{merity2018analysis}. A quasi-recurrent layer comprises two 
separate parts:\ a convolution layer with three weights, and a recurrent pooling 
layer. Given an input $\mathbf{X} \in \mathbb{R}^{k \times n}$, the convolution 
layer is
\begin{align*}
\mathbf{Z} = \tanh(\WW_z \cdot \mathbf{X})\\
\mathbf{F} = \sigma(\WW_f \cdot \mathbf{X})\\
\mathbf{O} = \sigma(\WW_o \cdot \mathbf{X})
\end{align*}
\noindent where $\sigma$ denotes the sigmoid function, $\cdot$ represents 
masked convolution across time, and $\WW_{\{z, f, o\}} \in \mathbb{R}^{m \times 
k \times r}$ are convolution weights with $k$ input channels, $m$ output 
channels, and a window size of $r$. In the recurrent pooling layer, the 
convolution outputs are combined sequentially:
\begin{align*}
\mathbf{c}_t &= \mathbf{f}_t \odot \mathbf{c}_{t-1} + (1 - 
\mathbf{f}_t) \odot \mathbf{z}_t\\
\mathbf{h}_t &= \mathbf{o}_t \odot \mathbf{c}_t
\end{align*}
\noindent Multiple QRNN layers can be stacked for deeper hierarchical 
representation, with the output $\mathbf{h}_{1:t}$ being fed as the input into 
the subsequent layer: In language modeling, a four-layer QRNN is a 
standard architecture~\cite{merity2018analysis}.

\parheader{Perplexity--Recall Scale}\label{section:pplr}
Word-level perplexity does not have a strictly monotonic relationship with 
recall-at-$k$, the fraction of top $k$ predictions that contain the correct word.
A given R@$k$ imposes a weak minimum perplexity 
constraint---there are many free parameters that allow for large variability in 
the perplexity given a certain R@$k$. Consider the corpus, ``choo 
choo train,'' with an associated unigram model $P(\text{``choo''}) = 0.1$, 
$P(\text{``train''}) = 0.9$, resulting in an R@1 of $1/3$ and perplexity of 
$4.8$. Clearly, R@1 $ =1/3$ for 
all $P(\text{``choo''}) \leq 0.5$; thus, perplexity can drop as low as $2$ 
without affecting recall.

\section{Experimental Setup}

We conducted our experiments on Penn Treebank~(PTB; \citealp{ptb}) and 
WikiText-103~(WT103; \citealp{wikitext}). Preprocessed by \citet{mikolov}, PTB 
contains 887K tokens for training, 70K for validation, and 78K for test, with a 
vocabulary size of 10,000. On the other hand, WT103 comprises 103 million 
tokens for training, 217K for validation, and 245K for test, spanning a 
vocabulary of 267K unique tokens.

For the neural language model, we used a four-layer 
QRNN~\cite{bradbury2016quasi}, which achieves state-of-the-art results on a 
variety of datasets, such as WT103~\cite{merity2018analysis} and PTB. 
To compare against more common LSTM architectures, we also evaluated 
AWD-LSTM~\cite{merity2017regularizing} on PTB. For the non-neural approach, we 
used a standard five-gram model with modified Kneser-Ney 
smoothing~\cite{chen1996empirical}, as explored in \citet{mikolov2012context} 
on PTB. We denote the QRNN models for PTB and WT103 as {\tt ptb-qrnn} and {\tt 
wt103-qrnn}, respectively.

For each model, we examined word-level perplexity, R@3 in next-word prediction, latency (ms/q), 
and energy usage (mJ/q). To explore the perplexity--recall 
relationship, we collected individual perplexity and recall statistics for each 
sentence in the test set.

\subsection{Hyperparameters and Training}

The QRNN models followed the exact training procedure and architecture 
delineated in the official codebase 
 from \citet{merity2018analysis}. For {\tt ptb-qrnn}, we trained the model for 
 550 epochs using NT-ASGD~\cite{merity2017regularizing}, then finetuned for 300 
 epochs using 
 ASGD~\cite{asgd}, all with a learning rate of 30 throughout. For {\tt 
 wt103-qrnn}, we followed \citet{merity2018analysis} and trained the QRNN for 
 14 epochs, using the Adam optimizer with a learning rate of $10^{-3}$. We also 
 applied regularization techniques from \citet{merity2017regularizing}; all the 
 specific hyperparameters are the same as those in the repository. Our model 
 architecture consists of 400-dimensional tied embedding 
 weights~\cite{inan2016tying} and four QRNN layers, with 1550 hidden units per 
 layer on PTB and 2500 per layer on WT103. Both QRNN models have window sizes 
 of $r=2$ for the first layer and $r=1$ for the rest.
 
For the KN-5 model, we trained an off-the-shelf five-gram model using the 
popular SRILM toolkit~\cite{stolcke2002srilm}. We did not specify any special 
hyperparameters.
 
\subsection{Infrastructure}
We trained the QRNNs with PyTorch (0.4.0; commit {\tt 1807bac}) on a 
 Titan V GPU. To evaluate the models under a 
 resource-constrained environment, we deployed them 
on a Raspberry Pi 3 (Model B) running Raspbian Stretch (4.9.41-v7+). The 
Raspberry Pi (RPi) is not only a standard platform, but also a 
close surrogate to mobile phones, using the same Cortex-A7 in many phones.
We then transferred the trained models to the RPi, using the same frameworks 
for evaluation. We plugged the RPi into a Watts Up Pro meter, a power meter 
that can be read programatically over USB at a frequency of 1 Hz. For the 
QRNNs, we used the first 350 words of the test set, and averaged the ms/query 
and mJ/query. For KN-5, we used the entire test set for evaluation, since the 
latency was much lower. To adjust for the base power load, we subtracted idle 
power draw from energy usage.

For a different perspective, we further evaluated all the 
models under a desktop environment, using an i7-4790k CPU and Titan V GPU. 
Because the base power load for powering a desktop is much higher than
running neural language models, we collected only latency statistics. We used 
the entire test set, since the QRNN runs quickly.

In addition to energy and latency, another consideration for the NLP developer 
selecting an operating point is the cost of underlying hardware. For our setup, 
the RPi costs \$35 USD, the CPU costs \$350 
USD, and the GPU costs \$3000 USD.

\section{Results and Discussion}
\begin{table}[t]
  \centering
  \setlength{\tabcolsep}{3pt}
  \begin{tabular}{l c c}
    \toprule[1pt]
    {\bf Model} & {\bf Val.} & {\bf Test}\\
    \midrule
    \addlinespace[-0.5pt]
    \multicolumn{3}{c}{{\bf\small Penn Treebank}}\\
    \addlinespace[-3pt]
    \midrule
    Skip LSTM~\cite{melis2017state} & 60.9 & 58.3\\
    AWD-LSTM~\cite{merity2017regularizing} & 60.0 & 57.3\\
    QRNN & 59.1 & 56.8\\
    \midrule
    \addlinespace[-0.5pt]
    \multicolumn{3}{c}{{\bf\small WikiText-103}}\\
    \addlinespace[-3pt]
    \midrule
    Rae-LSTM~\cite{rae2018fast} & 36.0 & 36.4\\
    QRNN & 31.9 & 32.8\\
    \bottomrule[1pt]
  \end{tabular}
  \caption{Comparison of neural language models on Penn Treebank and 
    WikiText-103.}
  \label{table:stoa}\vspace{-5mm}
\end{table}

\begin{figure*}[t]
  \centering\vspace{-5mm}
  \begin{minipage}{0.48\linewidth}
    \centering
    \subfloat{\stackunder[3pt]{\includegraphics[scale=0.3]{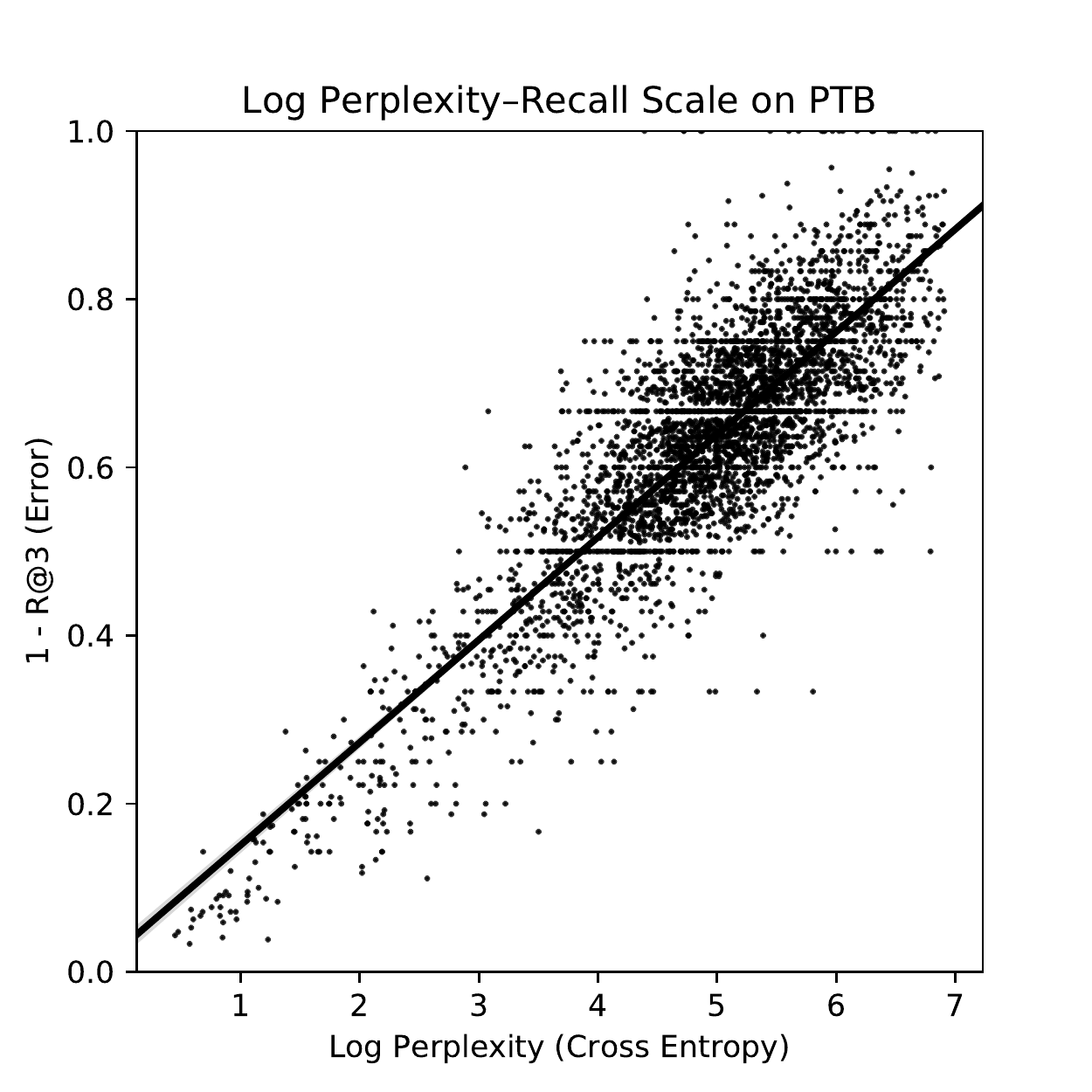}}{$r^2=0.72$}}
    \subfloat{\stackunder[3pt]{\includegraphics[scale=0.3]{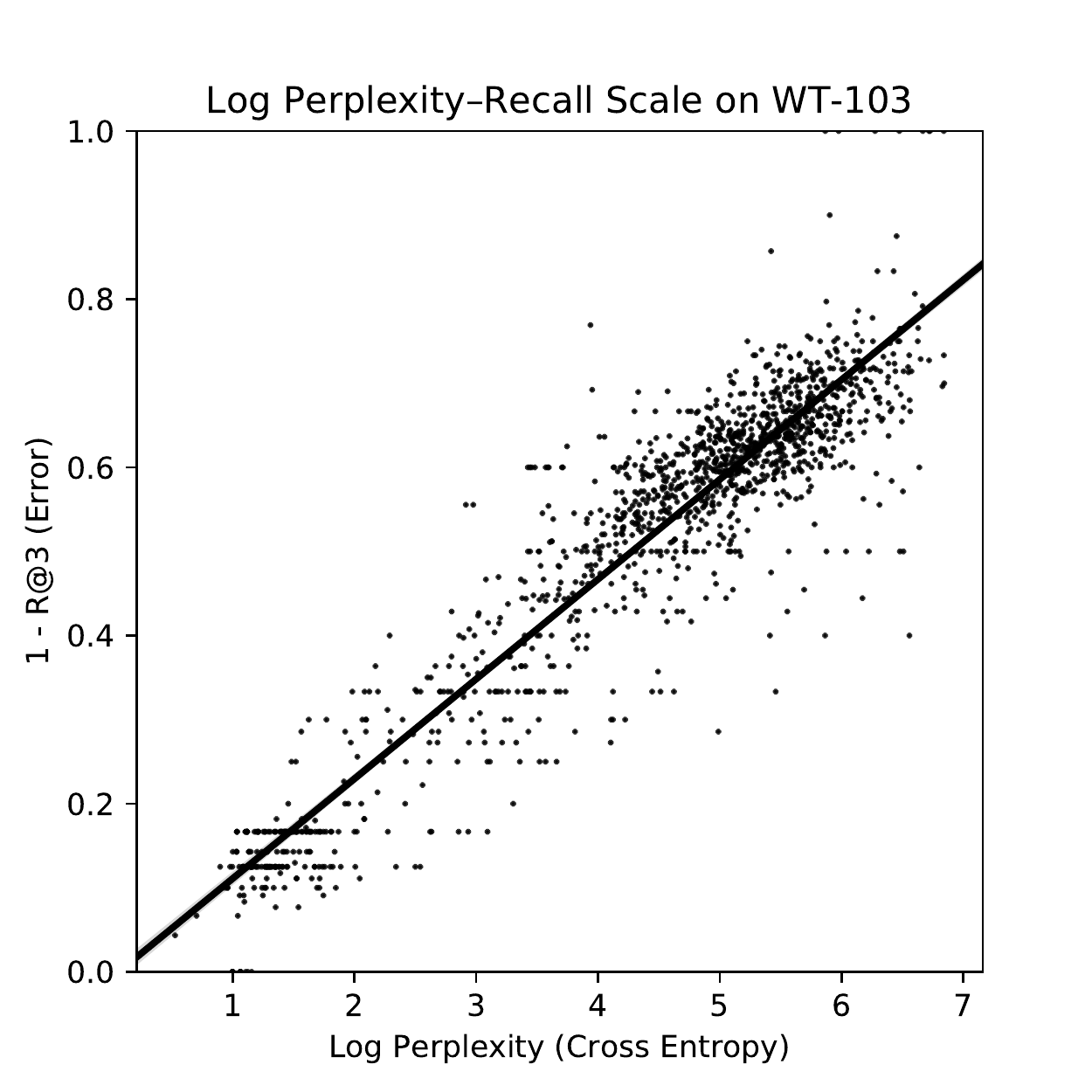}}{$r^2=0.88$}}
    \caption{Log perplexity--recall error with KN-5.}
    \label{fig:kn}
  \end{minipage}
  \begin{minipage}{0.48\linewidth}
    \centering
    \subfloat{\stackunder[3pt]{\includegraphics[scale=0.3]{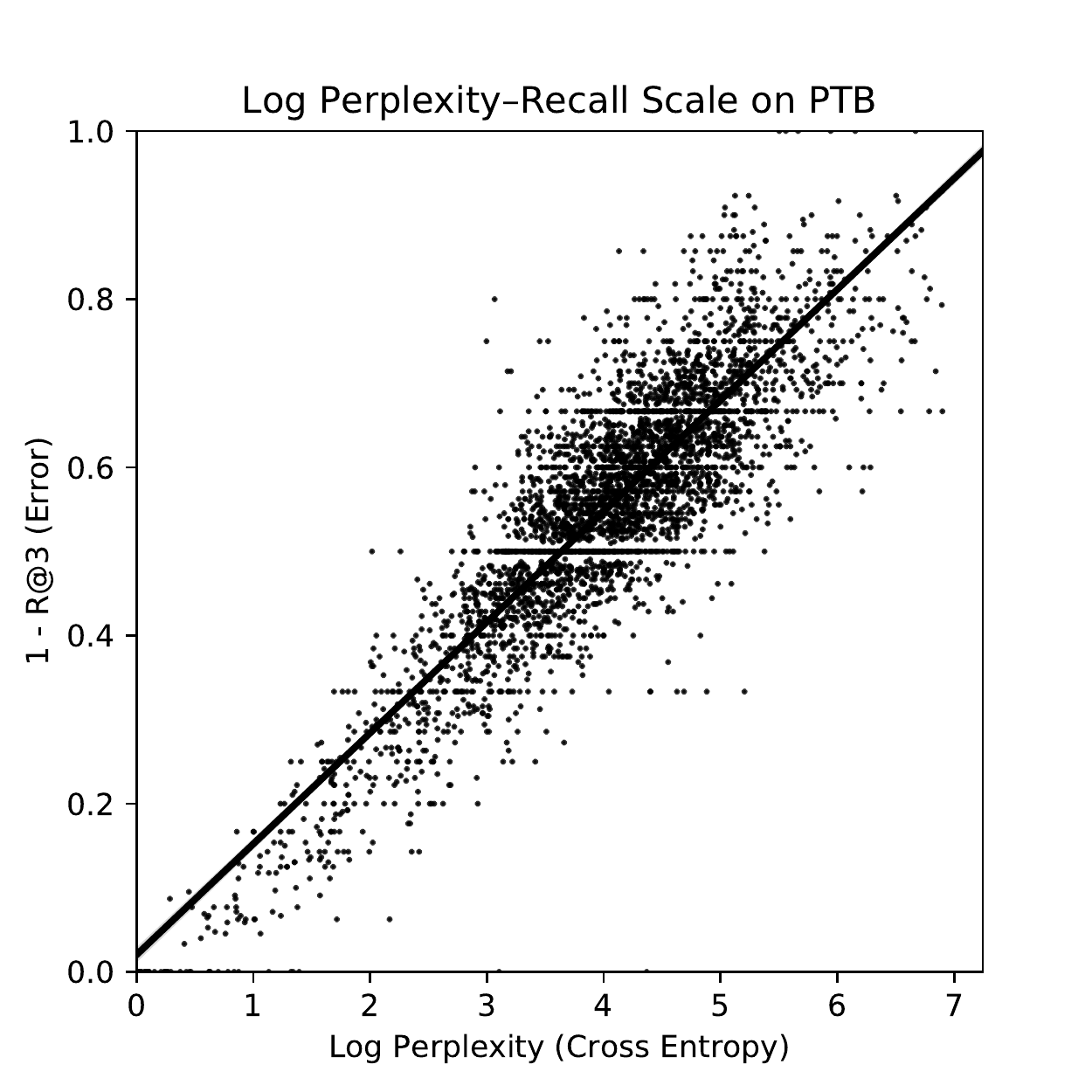}}{$r^2=0.77$}}
    \subfloat{\stackunder[3pt]{\includegraphics[scale=0.3]{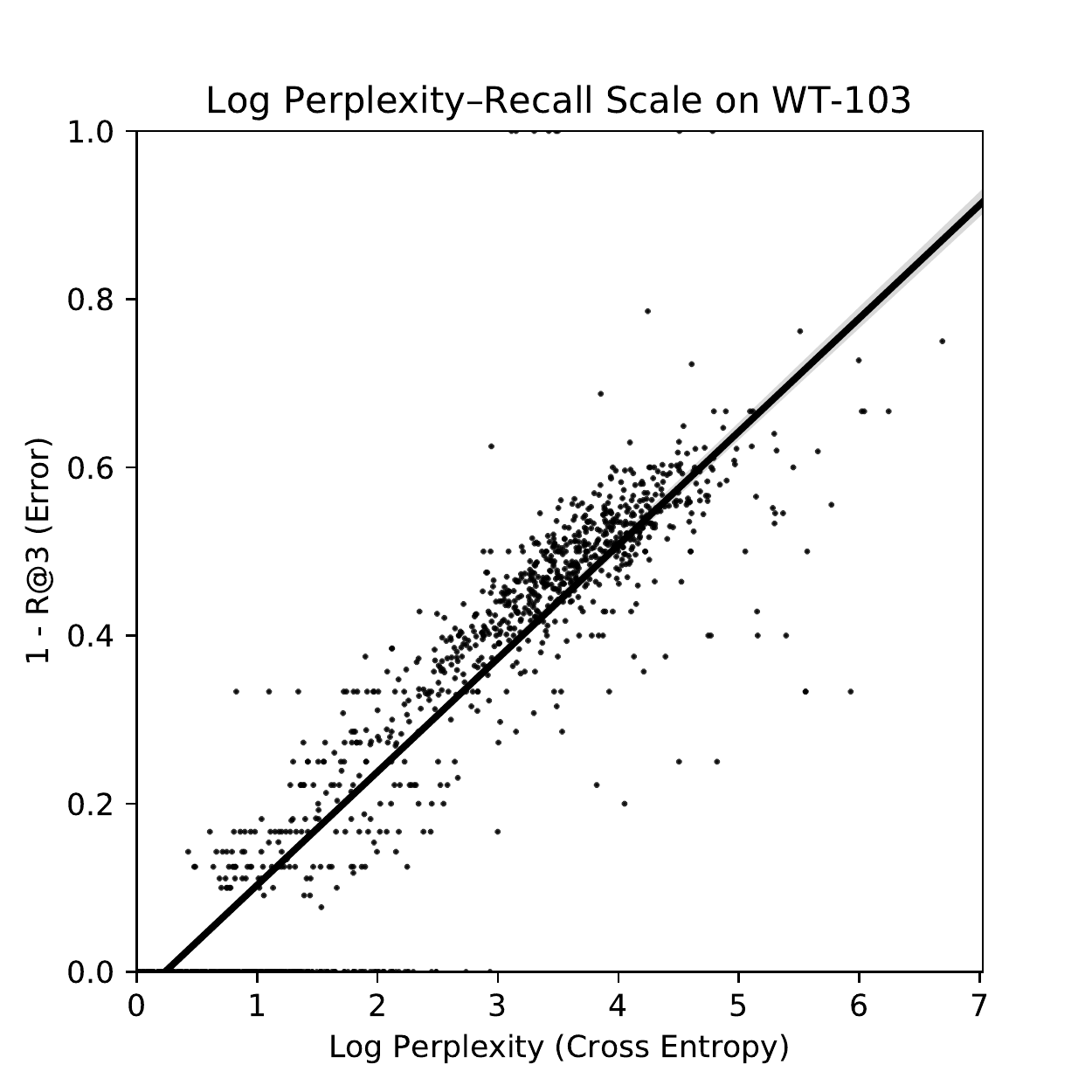}}{$r^2=0.86$}}
    \caption{Log perplexity--recall error with QRNN.}
    \label{fig:qrnn}
  \end{minipage}\vspace{-3mm}
\end{figure*}

To demonstrate the effectiveness of the QRNN models, we present the 
results of past and current state-of-the-art neural language models in 
Table~\ref{table:stoa}; we report the Skip- and AWD-LSTM results as seen 
in the original papers, while we report our QRNN results.
 Skip LSTM denotes the four-layer Skip LSTM in 
\citet{melis2017state}. \citet{rae2018fast} focus on Hebbian softmax, a model 
extension technique---Rae-LSTM refers to their base LSTM model 
without any extensions. In our results, KN-5 refers to the traditional 
five-gram model with modified Kneser-Ney smoothing, and AWD is shorthand for 
AWD-LSTM. 

\parheader{Perplexity--recall scale}
In Figure \ref{fig:kn}, using KN-5 as the model, we plot the log perplexity 
(cross entropy) and R@3 error ($1 - \text{R@3}$) for every sentence in PTB and 
WT103. 
The horizontal clusters arise from multiple 
perplexity points representing the same R@3 value, as explained in Section 
\ref{section:pplr}. We also observe that the perplexity--recall scale is 
non-linear---instead, log perplexity appears to have a moderate linear 
relationship with R@3 error on PTB ($r=0.85$), and an even stronger 
relationship on WT103 ($r=0.94$). This is partially explained by WT103 having 
much longer sentences, and thus less noisy statistics.

From Figure \ref{fig:qrnn}, we find that QRNN models yield strongly linear log 
perplexity--recall plots as well, where $r=0.88$ and $r=0.93$ for PTB and 
WT103, respectively. Note that, due to the improved model quality over KN-5, 
the point clouds are shifted downward compared to Figure \ref{fig:kn}. We 
conclude that log perplexity, or 
cross entropy, provides a more human-understandable indicator of R@3 than 
perplexity does. Overall, these findings agree with those from 
\citet{chen1998evaluation}, which explores the log perplexity--word error rate
scale in language modeling for speech recognition.
\begin{table}[t]
  \centering
  \small
  \setlength{\tabcolsep}{2pt}
  \begin{tabular}{c l c  c  c  c  c c c }
    
    \toprule[1pt]
    \multirow{2}{*}{\raisebox{-3\heavyrulewidth}{\bf \#}} &
    \multirow{2}{*}{\raisebox{-3\heavyrulewidth}{\bf Method}} & 
    \multicolumn{3}{c}{\bf\small Model Quality} &
    \multicolumn{2}{c}{\bf\small RPi} &
    \multicolumn{2}{c}{\bf\small CPU $|$ GPU}\\
    \cmidrule(lr){3-5} 
    \cmidrule(lr){6-7}
    \cmidrule(lr){8-9}& &
    Val. &  Test &  R@3 & 
    ms/q & 
    mJ/q& 
    ms/q & 
    ms/q \\
    \midrule
    \multicolumn{9}{c}{{\bf\small Penn Treebank}}\\
    \addlinespace[-1.6pt]
    \midrule
    1 & KN-5 & 148.4 & 141.5 & 36.7\% & 7 & 6 & 0.8 & -- \\
    2 & AWD & 59.2 & 56.8 & 44.9\% & 223 & 295 & 7.9 & 1.7 \\
    3 & QRNN & 59.1 & 56.8 & 44.7\% & 224 & 296 & 7.5 & 1.6 \\
    \midrule
    \multicolumn{9}{c}{{\bf\small WikiText-103}}\\
    \addlinespace[-1.6pt]
    \midrule
    4 & KN-5 & 145.2 & 152.7 & 39.8\% & 264 & 229 & 37 & -- \\
    5 & QRNN & 31.9 & 32.8 & 53.5\% & 1240 & 1480 & 59 & 3.5 \\
    \bottomrule[1pt]
  \end{tabular}
  \caption{Language modeling results on performance and model quality.}
  \label{table:ptb}\vspace{-5mm}
\end{table}

\parheader{Quality--performance tradeoff}
In Table \ref{table:ptb}, from left to right, 
we report perplexity results on the validation and test sets, R@3 on test, and finally 
per-query latency and energy usage. On the RPi, KN-5 is both fast and power-efficient to run, 
using only about 7 ms/query and 6 mJ/query for PTB (Table \ref{table:ptb}, row 1), and 264 
ms/q and 229 mJ/q on WT103 (row 5). Taking 220 
ms/query and consuming 300 mJ/query, AWD-LSTM and {\tt ptb-qrnn} are still 
viable for mobile phones: The modern smartphone holds upwards of 10,000 
joules~\cite{carroll2010analysis}, and the latency is within usability 
standards~\cite{miller1968}. Nevertheless, the models are still 49$\times$ 
slower and 32$\times$ more power-hungry than KN-5. The {\tt wt103-qrnn} model is 
completely unusable on phones, taking over 1.2 seconds per next-word 
prediction. Neural models achieve perplexity drops of 60--80\% and R@3 
increases of 22--34\%, but these improvements come at a much higher cost in 
latency and energy usage.

In Table \ref{table:ptb} (last two columns), the desktop yields very 
different results: the neural models on PTB (rows 2--3) are 9$\times$ slower 
than KN-5, but the absolute latency is only 8 ms/q, which is still much faster 
than what humans perceive as instantaneous~\cite{miller1968}. If a high-end 
commodity GPU 
is available, then the models are only twice as slow as KN-5 is. From row 5, 
even better results are noted with {\tt wt103-qrnn}: On the CPU, the QRNN is 
only 60\% slower than KN-5 is, while the 
model is faster by 11$\times$ on a GPU. These results suggest that, if only 
latency is considered under a commodity desktop environment, the QRNN model is 
humanly indistinguishable from the KN-5 model, even without using GPU 
acceleration.

\section{Conclusion}

In the present work, we describe and examine the tradeoff space between quality 
and performance for the task of language modeling. Specifically, we explore the 
quality--performance tradeoffs between KN-5, a non-neural approach, and 
AWD-LSTM and QRNN, two neural language models. We find that with decreased 
perplexity comes vastly increased computational requirements: In one of the 
NLMs, a perplexity reduction by 2.5$\times$ results in a 49$\times$ rise in 
latency and 32$\times$ increase in energy usage, when compared to KN-5.

\end{document}